\documentclass[twoside,leqno,twocolumn]{article}

\usepackage[letterpaper]{geometry}
\usepackage{booktabs}
\usepackage{ltexpprt}
\usepackage{hyperref}
\usepackage{balance}
\newtheorem{assumption}{Assumption}[section]
\newcommand{\independent}{\protect\mathpalette{\protect\independenT}{\perp}}
\def\independenT#1#2{\mathrel{\rlap{$#1#2$}\mkern2mu{#1#2}}}
\usepackage{graphicx}
\usepackage{multirow}
\usepackage{amssymb}
\usepackage{pifont}
\usepackage{amsmath}
\usepackage{diagbox}
\usepackage{url}
\usepackage{hyperref}
\usepackage{longtable}
\usepackage{todonotes}
\usepackage{amsmath}
\usepackage{wrapfig}
\usepackage{algorithm,algcompatible,amsmath}

\algnewcommand\INPUT{\item[\textbf{Input:}]}
\algnewcommand\OUTPUT{\item[\textbf{Output:}]}

\begin{document}

\title{\Large Learning Infomax and Domain-Independent Representations for Causal Effect Inference with Real-World Data}
\author{Zhixuan Chu\thanks{Ant Group (chuzhixuan.czx@alibaba-inc.com).}
\and Stephen L. Rathbun\thanks{University of Georgia (rathbun@uga.edu).}
\and Sheng Li\thanks{University of Georgia (sheng.li@uga.edu).}}

\date{}

\maketitle

\fancyfoot[R]{\scriptsize{Copyright \textcopyright\ 2022 by SIAM\\
SIAM International Conference on Data Mining (SDM22)}}

\begin{abstract} \small\baselineskip=9pt

The foremost challenge to causal inference with real-world data is to handle the imbalance in the covariates with respect to different treatment options, caused by treatment selection bias. To address this issue, recent literature has explored domain-invariant representation learning based on different domain divergence metrics (e.g., Wasserstein distance, maximum mean discrepancy, position-dependent metric, and domain overlap). In this paper, we reveal the weaknesses of these strategies, i.e., they lead to the loss of predictive information when enforcing the domain invariance; and the treatment effect estimation performance is unstable, which heavily relies on the characteristics of the domain distributions and the choice of domain divergence metrics. Motivated by information theory, we propose to learn the Infomax and Domain-Independent Representations to solve the above puzzles. Our method utilizes the mutual information between the global feature representations and individual feature representations, and the mutual information between feature representations and treatment assignment predictions, in order to maximally capture the common predictive information for both treatment and control groups. Moreover, our method filters out the influence of instrumental and irrelevant variables, and thus it effectively increases the predictive ability of potential outcomes.  Experimental results on both the synthetic and real-world datasets show that our method achieves state-of-the-art performance on causal effect inference. Moreover, our method exhibits reliable prediction performances when facing data with different characteristics of data distributions, complicated variable types, and severe covariate imbalance.

\end{abstract}


\section{Introduction}

Real-world data refer to observational data as opposed to data gathered in an experimental setting such as a randomized controlled trial (RCT). They are derived from electronic health records (EHRs), claims and billing activities, product and disease registries, etc \cite{bian2020assessing}. Due to the hallmark of real-world data, subjects would have a preference for a certain treatment option, which leads to a bias of the distribution for the covariates among different treatment options. The selection bias makes the distribution of the covariates in the treatment group different from the control group, and such a huge discrepancy between the treatment and control groups exacerbates the difficulty of counterfactual outcome estimation.  Therefore, how to handle the selection bias is a challenging problem in causal effect estimation.

Recent causal effect estimation methods~\cite{johansson2016learning, shalit2017estimating,li2017matching} have built a strong connection with domain adaptation, by enforcing domain invariance with distributional distances such as the Wasserstein distance and maximum mean discrepancy. Inspired by metric learning, some methods~\cite{yao2018representation} use hard samples to learn representations that preserve local similarity information and balance the data distributions. In~\cite{zhang2020learning}, the authors argue that distribution invariance is often too strict a requirement and they propose to use counterfactual variance to measure the domain overlap. Thus, for the domain adaptation problem under causal inference settings, which is the best measurement for the imbalanced domains remains unsettled and is highly relies on the characteristics of the distributions of domains and the hyperparameter of regularization term for imbalance mitigation~\cite{yao2020survey}. Besides, despite the empirical success of such methods, enforcing balance can, to various extents, remove predictive information and lead to a loss in predictive power, regardless of which type of domain divergence metric is employed~\cite{alaa2018limits}. 

Besides, when handling the selection bias, there is another issue that can lead to poor potential outcome estimation, i.e., the types of observed variables. The major drawback of existing causal inference methods is that they always treat all observed variables as pre-treatment variables, which are not affected by treatment assignments but may be predictive of outcomes. This assumption is not tenable for observational data. If all observed variables are directly used to estimate treatment effects, more impalpable bias may be introduced into the model. For example, conditioning on an instrumental variable, which is associated with the treatment assignment but not with the outcome except through exposure, can increase both bias and variance of estimated treatment effects~\cite{myers2011effects_instrumental}. 

To sum up, successfully estimating the causal effect needs three desiderata, i.e., filtering out information about instrumental and irrelevant variables, capturing the predictive information, and mitigating the covariate imbalance between treatment and control groups. To achieve these three desiderata simultaneously, we propose an Infomax and Domain-independent Representation Learning (IDRL) method to estimate the causal effects with observational data by seeking a representation space, which not only contains the common predictive information about potential outcome estimation but also excludes the domain-dependent information. IDRL relies on two mutual information structures: one is to maximize the mutual information between global summary representation and individual feature representation, which can maximally capture the common predictive information for both treatment and control groups, and filter out the noise only for specific individual or group; the other is to minimize the mutual information between feature representation vector and treatment options, which makes feature representations independent from treatment option domains. Therefore, instead of enforcing balance between the treatment and control groups by adopting various domain divergence metrics, our IDRL method utilizes one mutual information module to exclude the information related to the domain, so that we cannot tell which domain it is from. At the same time, another mutual information can maximally preserve common predictive information. 

Our main contributions are summarized in the following: $(1)$ our work is the first attempt to utilize the global summary representations for causal effect estimation, which help capture the common predictive information for both treatment and control groups. $(2)$ Circumventing the strategy of enforcing balance between treatment and control groups, our IDRL method learns the domain-independent representation to solve the selection imbalance problem. $(3)$ Our method achieves state-of-the-art performance on treatment effect estimation benchmarks such as IHDP, Jobs, and News.

\section{Background}

Let $t_i$ denote the treatment assignment for unit $i$; $i=1,...,n$. In the binary treatment case, unit $i$ will be assigned to the treatment group if $t_i = 1$, or to the control group if $t_i=0$. Let $X \in \mathbb{R}^d$ denote all observed variables of a unit. In the real-world data, only the factual outcomes are available, while the counterfactual outcomes have never been observed. The potential outcome for unit $i$ is denoted by $Y_{t}^i$ when treatment $t$ is applied to unit $i$. Then the observational data can be denoted as $\{x_i, t_i, y_i\}_{i=1}^n$. The individual treatment effect (ITE) for unit $i$ is the difference between the potential treated and control outcomes, which is defined as: 
\begin{equation}
    \text{ITE}_i = Y_1^i - Y_0^i, \quad (i=1,...,n).
\end{equation}

The average treatment effect (ATE) is the difference between the mean potential treated and control outcomes, which is defined as:

\begin{equation}
\text{ATE}=\frac{1}{n}\sum_{i=1}^{n}(Y_1^i - Y_0^i), \quad (i=1,...,n).
\end{equation}

Existing counterfactual estimation methods usually make the strong ignorability assumption, which ensures that the treatment effect can be identified in the potential outcome framework~\cite{rubin1974estimating,imbens2015causal}.

\begin{assumption} \textbf{Strong Ignorability}: 
Given covariates $X$, treatment assignment $T$ is independent of the potential outcomes, i.e., $(Y_1, Y_0) \independent T | X$ and for any value of $\,X$, treatment assignment is not deterministic, i.e., $P(T = t | X = x) > 0$, for all $t$ and $x$.
\end{assumption}

Strong ignorability is also known as the no unmeasured confounders assumption, which makes ITE identifiable. This assumption requires that the observed covariates are sufficient to characterize the treatment assignment mechanism. Besides, if for some values of $X$, the treatment assignment is deterministic, there is no observed outcome available in one of the treatment groups. In that case, it is meaningless to estimate the counterfactual outcomes.

\section{Proposed Framework}
\subsection{Motivation}

The most challenging issue in causal effect estimation from observational data is how to properly handle the covariate shift problem caused by treatment selection bias. Due to the selection bias, there is an imbalance between treatment and control groups. This phenomenon exacerbates the difficulty of counterfactual outcome estimation. In addition, the instrumental and irrelevant variables also can bring bias and variance into the counterfactual outcome estimation. To address the selection bias issue, we propose the \textbf{Infomax and Domain-Independent Representation  Learning  (IDRL)} method to preserve the highly common predictive information for potential outcomes. Our method abandons the traditional distribution balancing strategies because different domain divergence metrics are sensitive to the characteristics of data distributions and the predictive information may be inadvertently removed when enforcing the balance procedure.

\begin{figure}[t!]
    \centering
    \includegraphics[width=1\columnwidth]{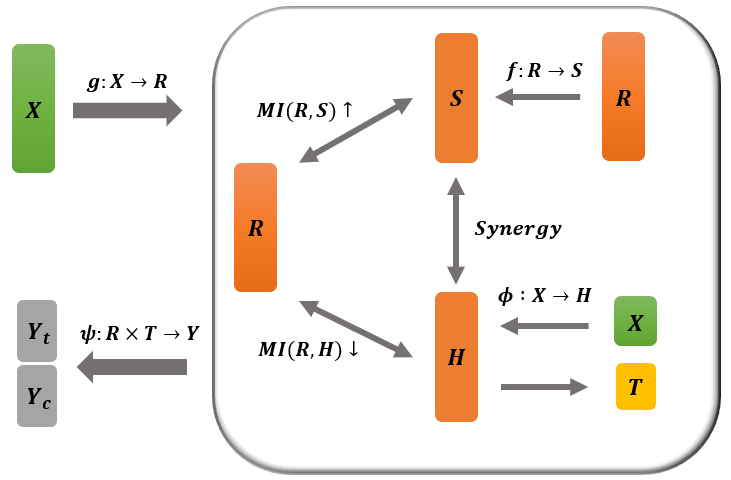}
    \caption{The framework of the proposed IDRL that consists of four main components, including feature representation learning $g : X \rightarrow R$, information maximization learning $MI(R,S)$, domain-independent learning $MI(R,H)$, and potential outcome generator $\psi: R \times T \rightarrow Y$. IDRL first learns an individual representation vector for each subject. At the same time, the information maximization learning and domain-independent learning are incorporated into the representation learning procedure to filter out domain-dependent information, solve the selection bias, and preserve the common predictive information for treatment and control groups.}
    \label{fig: framework}
\end{figure}

\subsection{Model Architecture}
As shown in Fig.~\ref{fig: framework}, our IDRL method consists of four main components, including feature representation learning, information maximization learning, domain-independent learning, and potential outcome generator. In the feature representation learning, IDRL first learns an individual representation vector for each subject via the standard feed-forward deep neural network. At the same time, the information maximization learning and domain-independent learning are incorporated into the representation learning procedure to filter out domain-dependent information, solve the selection bias, and preserve the common predictive information for treatment and control groups. Finally, the potential outcomes can be inferred by the outcome generator based on the learned representation. In the following, we present the details of each component.

\textbf{Feature Representation Learning.}
This step is to learn the feature representations of observed covariates by a function $g : X \rightarrow R, R \in \mathbb{R}^d$, which is parameterized by a deep neural network. The function $g(\cdot)$ maps the original covariate space $X$ into a $d$-dimensional representation space $R =\{r_1, r_2,...,r_n\}$. 

\textbf{Information Maximization Learning.}
Inspired by a recent unsupervised representation learning method that exploits individual and global information~\cite{hjelm2018learning, velickovic2019deep,chu2021graph}, we maximize the mutual information between the individual representation and global representation, such that the representation space could capture the common predictive information for treatment and control groups. Specifically, we utilize a summary function, $f : R^{n\times d} \rightarrow S, S \in \mathbb{R}^d$, which summarizes the learned individual representation into a global representation vector, i.e., $s=f(g(X))$. From the observations in empirical evaluations, the summary function could be defined as weighted averaging of all the subjects' representations: $s = \sigma(\frac{1}{2n_t}\sum_{i\in n_t}r_i+\frac{1}{2n_c}\sum_{i\in n_c}r_i)$ to best capture the global representation, where $\sigma$ is the logistic sigmoid activation function, $n_t$ and $n_c$ are the subject numbers of treatment and control groups, respectively. Our goal is to make all the subjects' representations preserve the common predictive information used to estimate the potential outcomes for treatment and control groups. Therefore, we aim at maximizing the mutual information $MI(r_i,s)$ between the learned individual representation $r_i$ and global summary representation $s$, where the feature representation learning can pick and choose what type of information in the original covariates is preserved into the learned representation vector. If the feature representation learning passes information specific to only some individuals or certain treatment options, this does not increase the mutual information with any of the other subjects. This encourages the feature representation learning to prefer information that is shared across the subjects. More details about the mutual information $MI(r_i,s)$ will be discussed later.

\textbf{Domain-Independent Learning.} To handle the selection bias, we incorporate a domain-independent learning module, which helps the subject's feature representation to be independent of its domain, instead of enforcing the domain invariance by various metrics (e.g., Wasserstein distance, maximum mean discrepancy). When the feature representations are independent of the domains, we cannot tell which domain the subject is from and thus filter out the information about the treatment assignments. Because the mutual information is small when the two variables are statistically independent, while it is large when two variables preserve the same information content, we employ the mutual information to measure the independence between feature representations and domains. To give full expression to the treatment domain information, we utilize the treatment domain prediction $H$ to represent the treatment domain by function $\phi: X \rightarrow H \rightarrow T$, rather than directly using treatment domain indicator $T$. Therefore, we aim at minimizing the mutual information $MI(r_i,h_i)$ between learned representation space $r_i$ and treatment domain prediction $h_i$. 

\textbf{Potential Outcome Generator.} 
So far, we have learned the feature representation space from feature representation learning, along with information maximization learning and domain-independent learning. The function $\psi: R \times T \rightarrow Y$ maps the representation vectors as well as the treatment assignment to the corresponding potential outcome, which is parameterized by a feed-forward deep neural network with multiple hidden layers and non-linear activation functions. To avoid the risk of losing the influence of $T$ when the dimension of representation space is high, $\psi: R \times T \rightarrow Y$ is partitioned into two head layers for treatment and control groups, separately. The output of $\psi$ estimates potential outcomes across treatment and control groups, including the estimated factual outcome $\hat{y}^f$ and the estimated counterfactual outcomes $\hat{y}^{cf}$. The factual outcomes $y^f$ are used to minimize the loss of prediction $\hat{y}^f$. We aim to minimize the mean squared error in predicting factual outcomes:
\begin{equation}
\mathcal{L}_Y = \frac{1}{n}\sum_{i=1}^{N}(\hat{y}^f_i-y^f_i)^2,
\label{eq: factual}
\end{equation}
where $\hat{y}_i=\psi(r_i,t_i)$ denotes the inferred observed outcome of unit $i$ corresponding to the factual treatment $t_i$. 

\begin{figure}[t]
    \centering
    \includegraphics[width=0.95\columnwidth]{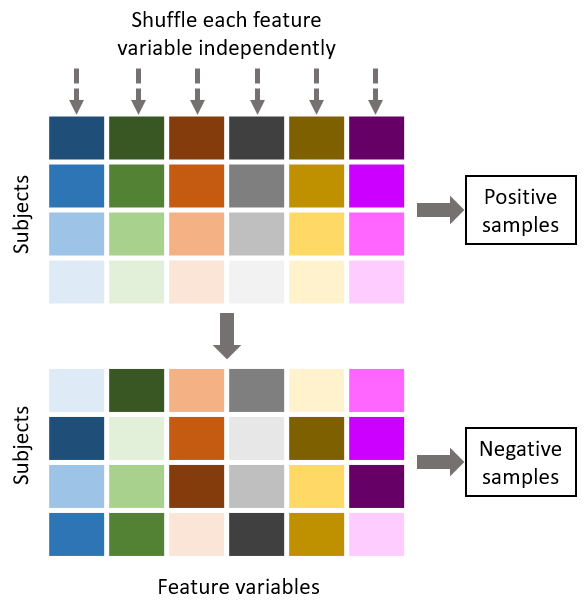}
    \caption{The strategy of generating negative samples where we independently shuffle feature variables of positive samples $X$ to generate negative samples $\Tilde{X}$. For example, there are 4 subjects (4 rows) and 6 feature variables (6 columns) and we independently shuffle each column of feature variable to generate negative samples.}
    \label{fig: shuffle}
\end{figure}

\textbf{Mutual Information Estimation.} In our model, it is unnecessary to use the exact KL-based formulation of MI, as we only want to maximize the mutual information $MI(r_i,s)$ between individual representation $r_i$ and global representation $s$, and minimize the mutual information  $MI(r_i,h_i)$ between individual representation $r_i$ and treatment domain prediction $h_i$. A simple and stable alternative based on the Jensen-Shannon divergence (JSD) can be utilized. Some recent work~\cite{belghazi2018mine,hjelm2018learning} has proved that an implicit estimation of mutual information can be attained with an encoder-discriminator architecture. Thus, we follow the intuitions from Deep Infomax~\cite{chu2021graph,hjelm2018learning} to optimize the mutual information involved in our method. To act as an agent for optimizing the mutual information, one discriminator is employed, which relies on a sampling strategy that draws positive and negative samples from the joint distribution and from the marginal product, respectively. To implement the discriminator, we also need to create the negative samples compared with original samples and then use the discriminator to distinguish which one is from positive samples (original data) and which one is from the negative samples (created fake data). The choice of the negative sampling procedure will govern the specific kinds of information that is desirable to be captured~\cite{velickovic2019deep}. Under causal inference settings, our main challenge is the covariate shift caused by selection bias, so we independently shuffle feature variables of positive samples $X$ to generate negative samples $\Tilde{X}$ as shown in Fig.~\ref{fig: shuffle}, which can break the imbalanced feature variable patterns in original $X$. Thus, for each feature variable in $\Tilde{X}$, there is no imbalance with respect to treatment options. 

For $MI(r_i,s)$,  one discriminator $d_s: R \times S \rightarrow P, P \in \mathbb{R}$ is employed. The discriminator is formulated by a simple bilinear scoring function with nonlinear activation: $d_s(r_i,s)=\sigma({r_i}^TWs)$, which estimates the probability of the $i$-th subject representation contained within the global representation $s$. $W$ is a learnable scoring matrix. We also conduct the feature representation learning for the negative samples $\tilde{X}$ to get the $\Tilde{r}_i$. With the proposed discriminator, we could have $d_s(r_i,s)$ and $d_s(\Tilde{r}_i,s)$, which indicate the probabilities of containing the representations of the $i$-th positive sample and negative sample in the global summary representation, respectively. We optimize the discriminator $d_s$ to maximize mutual information between $r_i$ and $s$ based on the Jensen Shannon divergence via a noise-contrastive type objective with a standard binary cross-entropy (BCE) loss~\cite{velickovic2019deep,hjelm2018learning}. The $\mathcal{L}_{MI(r_i,s)} $ is defined as:
\begin{equation}
\frac{1}{2n}\Big(\sum_{i=1}^n \mathbb{E}_{X}[\text{log} \, d(r_i,s)]+\sum_{j=1}^n \mathbb{E}_{\Tilde{X}}[\text{log} \, (1-d(\Tilde{r}_j,s))]\Big). 
\label{eq: mutual information1}
\end{equation}

For $MI(r_i,h_i)$,  one discriminator $d_h: R \times H \rightarrow P, P \in \mathbb{R}$ is adopted. Our method aims at optimizing the discriminator $d_h$, i.e., minimizing the mutual information between learned representation space $r_i$ and treatment domain prediction $h_i$. Similarly, the discriminator is formulated by a simple bilinear scoring function with nonlinear activation: $d_h(r_i,h_i)=\sigma({r_i}^T W h_i)$, where $r_i$ is the $i$-th subject's representation learned in feature representation learning procedure and $h_i$ is the $i$-th subject's treatment domain prediction learned from function $\phi$. We also attain the treatment domain prediction $\Tilde{h}_i$ for negative samples $\tilde{X}$ by function $\phi$. Therefore, in discriminator $d_h$, we could have $d_h(r_i,h_i)$ and $d_h(r,\tilde{h}_i)$, so the $\mathcal{L}_{MI(r_i,h_i)}$ is defined as:
\begin{equation}
\begin{split}
\frac{1}{2n}\Big(\sum_{i=1}^n \mathbb{E}_{X}[\text{log} \, d(r_i,h_i)]+\sum_{j=1}^n \mathbb{E}_{\Tilde{X}}[\text{log} \, (1-d(r_j,\Tilde{h}_j))]\Big).
\label{eq: mutual information2}
\end{split}
\end{equation}

\subsection{Overview of IDRL}

\begin{figure}
    \centering
    \includegraphics[width=0.95\columnwidth]{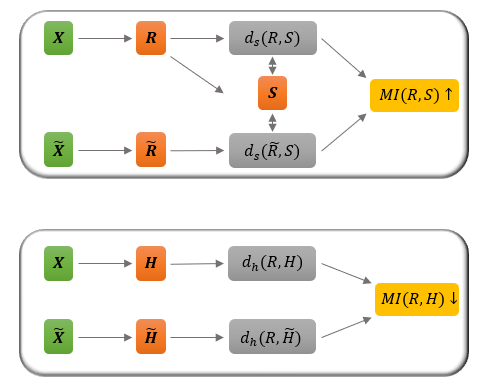}
    \caption{The procedures of IDRL. The top figure is information maximization learning that is used to maximize the mutual information between the individual representation ({$R$}) and global representation ($S$). The bottom figure is domain-independent learning that is used to help the subject’s feature representation ($R$) to be independent of its domain (treatment domain prediction $H$).}
    \label{fig: mi}
\end{figure}

The proposed IDRL method leverages the synergy between two mutual information modules to filter out the treatment domain information and noise, and thus capture the common predictive information for both treatment and control groups. In this way, our method can effectively increase the capability of predicting potential outcomes. As shown in Fig.~\ref{fig: mi}, we summarize the procedures of IDRL as follows:
\begin{enumerate}
\item  Create the negative samples $\tilde{X}$ by independently shuffling feature variables of positive samples $X$. 
\item  Learn the representation space $R$ for the positive samples $X$ and $\tilde{R}$  for the negative samples $\tilde{X}$ by function $g : X \rightarrow R$ and $g : \tilde{X} \rightarrow \tilde{R}$, respectively.
\item  Learn the treatment domain prediction $H$ for the positive samples $(X,T)$ by function $\phi : X \rightarrow H \rightarrow T$ and $\tilde{H}$ for the negative samples by plugging $\tilde{X}$ into function $\phi$. 
\item  Utilize a summary function $f : R^{n\times d} \rightarrow S$ to summarize the learned representation into a global summary representation, i.e., $s=f(g(X))$.
\item  Employ the discriminator $d_s: R \times S \rightarrow P$ to obtain $d_s(r_i,s)$ and $d_s(\tilde{r}_i,s)$, and the discriminator $d_h: R \times H \rightarrow P$ to obtain $d_h(h_i,r_i)$ and $d_h(\tilde{h}_i,r_i)$.
\item  Update parameters of $g$, $f$, $d_s$, and $d_h$ to maximize mutual information between $R$ and $S$ and minimize mutual information between $R$ and $H$, by applying gradient descent to maximize Eq.~(\ref{eq: mutual information1}) and minimize Eq.~(\ref{eq: mutual information2}).
\item  Use potential outcome generator $\psi: R \times T \rightarrow Y$ to estimate the potential outcomes by minimizing Eq.~(\ref{eq: factual})
\end{enumerate}

\section{Experiments}
In this section, we conduct experiments on three benchmarks, including the IHDP, Jobs, and News, to compare our proposed IDRL method with the state-of-the-art causal effect estimation methods. We also experiment with the synthetic datasets with different settings to validate the following aspects: (1) Capability of reliably predicting potential outcomes when facing data with different characteristics of the domain distributions and complicated variable types. (2) Robustness with respect to different levels of treatment selection bias. (3) The contribution of each component of the proposed IDRL method.

\subsection{Experiments on Benchmark Datasets}
\subsubsection{Datasets and Baseline Methods}

To evaluate our method and baselines on treatment effect estimation, we use the benchmarks, i.e., IHDP, Jobs, and News datasets. We compare our IDRL method with the following baseline methods: k-nearest neighbor \textbf{(kNN)}~\cite{ho2007matching}, 
Causal forests \textbf{(CF)}~\cite{wager2018estimation}, 
Random forest \textbf{(RF)}~\cite{breiman2001random}, 
Bayesian additive regression trees \textbf{(BART)}~\cite{chipman2010bart},  
Treatment-agnostic representation network \textbf{(TARNET)}~\cite{shalit2017estimating}, Counterfactual regression \textbf{($\text{CFRNET}_\text{wass}$)}~\cite{shalit2017estimating}, 
Local similarity preserved individual treatment effect estimation method \textbf{(SITE)}~\cite{yao2018representation}, 
Perfect match \textbf{(PM)}~\cite{schwab2018perfect}, and 
Causal Multi-task Gaussian Processes \textbf{(CMGP)}~\cite{alaa2017bayesian}. 

\subsubsection{Evaluation Metrics}
For IHDP and News datasets, both factual and counterfactual outcomes are generated from a known distribution, so that we are able to compute the expectations of the outcomes, the error of expected precision in estimation of heterogeneous effect ($\epsilon_\text{PEHE}$)~\cite{hill2011bayesian}, which is defined as:
\begin{equation}
    \epsilon_\text{PEHE}  = \frac{1}{n}\sum_{i=1}^{n}\Big(\mathbb{E}_{(y_1^i,y_0^i)	\sim \mu_Y(X)}[y_1^i-y_0^i]-[\hat{y}_1^i-\hat{y}_0^i] \Big)^2,
\end{equation}
where $y_1^i,y_0^i$ are treatment and control outcomes drawn from the ground truth $\mu_Y(X)$ and $\hat{y}_1^i,\hat{y}_0^i$ are estimated. The error of ATE estimation, which is defined as:
\begin{equation}
    \epsilon_\text{ATE}  = ||\frac{1}{n}\sum_{i=1}^{n}\mathbb{E}_{y^i	\sim \mu_Y(X)}[y^i]-\frac{1}{n}\sum_{i=1}^{n}\hat{y}^i||^2_2,
\end{equation}
where $\hat{y}^i$ is the estimated potential outcome.

For the Jobs dataset, because we only observe parts of the ground truth from the randomized experiment component, we cannot use the same evaluation metrics as IHDP. Therefore, we adopt the policy risk and ATT as~\cite{shalit2017estimating}:
\begin{equation}
\begin{split}
    R_{pol}(\pi)
    & = 1-[\mathbb{E}(y_1|\pi(x^i)=1)\cdot P(\pi=1)\\
    & +\mathbb{E}(y_0|\pi(x^i)=0)\cdot P(\pi=0)],
\end{split}
\end{equation}
where $\pi(x^i)=1$ when $\hat{y}_1^i-\hat{y}_0^i>0$.
\begin{equation}
\begin{split}
    &ATT=|T|^{-1}\sum_{i\in T}y^i-|C\cap E|^{-1}\sum_{i\in C\cap E}y^i,\\
    & \epsilon_\text{ATT} = |\text{ATT}-\frac{1}{|T|}\sum_{i\in T}(\hat{y}_1^i-\hat{y}_0^i)|,
\end{split}
\end{equation}
where all the treated subjects $T$ are part of the original randomized sample $E$, and $C$ is the control group.

For IHDP, Jobs, and News datasets, we report in-sample and out-of-sample performance.  In-sample is to estimate counterfactual outcomes for all units whose factual outcome of one treatment is observed. Out-of-sample is to estimate potential outcomes for units with no observed outcomes. This can help to select the best possible treatment for a new patient. In-sample error is computed over both the training and validation sets, and out-of-sample error over the test sets. 

\begin{table*}[h]
  \caption{Performance on IHDP, Jobs, and News. For IHDP and News, we present the mean value of $\sqrt{\epsilon_\text{PEHE}}$ and $\epsilon_\text{ATE}$ on in-sample and out-of-sample. For Jobs, we present the mean value of $R_{pol}$ and $\epsilon_\text{ATT}$ on in-sample and out-of-sample.}
  \label{result1}
  \centering
  \scalebox{0.9}{
  \begin{tabular}{lllllllllllll}
    \toprule
     & \multicolumn{4}{c}{IHDP} & \multicolumn{4}{c}{Jobs}&
     \multicolumn{4}{c}{News}
     \\
    \cmidrule(lr){2-5} \cmidrule(lr){6-9} \cmidrule(lr){10-13} 
    
     & \multicolumn{2}{c}{In-sample} & \multicolumn{2}{c}{Out-sample}  & \multicolumn{2}{c}{In-sample}  & \multicolumn{2}{c}{Out-sample}  
     & \multicolumn{2}{c}{In-sample}
     & \multicolumn{2}{c}{Out-sample}\\
    \cmidrule(lr){2-3} \cmidrule(lr){4-5} \cmidrule(lr){6-7} \cmidrule(lr){8-9} \cmidrule(lr){10-11} \cmidrule(lr){12-13} 
    Method     & $\sqrt{\epsilon_\text{PEHE}}$   & $\epsilon_\text{ATE}$  & $\sqrt{\epsilon_\text{PEHE}}$     & $\epsilon_\text{ATE}$ & 
    $R_{pol}$     & $\epsilon_\text{ATT}$ & $R_{pol}$   & $\epsilon_\text{ATT}$ 
    & $\sqrt{\epsilon_\text{PEHE}}$   & $\epsilon_\text{ATE}$
     & $\sqrt{\epsilon_\text{PEHE}}$   & $\epsilon_\text{ATE}$\\
    \midrule
    kNN &  2.1& 0.14 &4.1  & 0.79 & 0.23 & 0.02 & 0.26& 0.13 & 18.14& 7.83& 20.75& 9.14\\
    RF &  4.2& 0.73 & 6.6 & 0.96 & 0.23 & 0.03& 0.28 & 0.09 &17.39 & 5.5& 19.38&6.84\\
    CF &  3.8& 0.18 & 3.8 & 0.4 & 0.19 & 0.03 & 0.2 &  0.07 &17.59 &4.02 & 19.98&5.24\\
    BART & 2.1 & 0.23 & 2.3 & 0.34 & 0.23 & 0.02 & 0.25 & 0.08 & 18.53& 5.4& 21.04&6.73\\
    TARNET & 0.88 & 0.26 & 0.95 & 0.28 & 0.17 & 0.05 & 0.21 & 0.11 & 17.17&4.58 & 18.77&5.88\\
    CFRNET & 0.71 & 0.25 & 0.76 & 0.27 & 0.17 & 0.04 & 0.21 & 0.08 & 16.93& 4.54& 18.25&5.80\\
    SITE & 0.69 & 0.22 & 0.75 & 0.24 & 0.17 & 0.04 & 0.21 & 0.09 & 16.86 & 4.35& 18.20&5.56\\
    CMGP & \textbf{0.65} & \textbf{0.11} & 0.77 & \textbf{0.13} & 0.22 & 0.06 & 0.24 & 0.09 & 16.79 & 4.13& 17.89 &5.31\\
    PM & 0.68 & 0.20 & 0.74  & 0.22  &  0.16 & 0.03  & 0.18  & 0.05 & 16.76 & 3.99 &  17.80& 5.19\\
    \midrule
    IDRL (Ours) & 0.68 & 0.18 & \textbf{0.73} & 0.20 & \textbf{0.13} & \textbf{0.02} & \textbf{0.16} & \textbf{0.04} & \textbf{16.41} & \textbf{3.23} &\textbf{17.55} &\textbf{4.42}\\
    \bottomrule
  \end{tabular}
  }
\end{table*}

\subsubsection{Results and Analysis}  
Table \ref{result1} shows the performance of our method and baseline methods on the IHDP, Jobs, and News. Our method significantly outperforms all competing algorithms on the Jobs dataset and News datasets. On the IHDP dataset, our method has the best performance in the out-sample case and achieves comparable results with the best baselines, such as CFRNET, SITE, and CMGP in the in-sample case. The less-than-optimal in-sample results and more accurate out-sample results on IHDP more demonstrate our method's generalization of predictive power by capturing and preserving the common predictive information across all subjects.

\subsection{Experiments on Synthetic Datasets}

We further evaluate the performance of our method when facing data with different characteristics of the domain distributions, complicated variable types, and severe covariate imbalance. In addition, we evaluate the contribution of each component in our method.

\begin{figure}
    \centering
    \includegraphics[width=0.9\columnwidth]{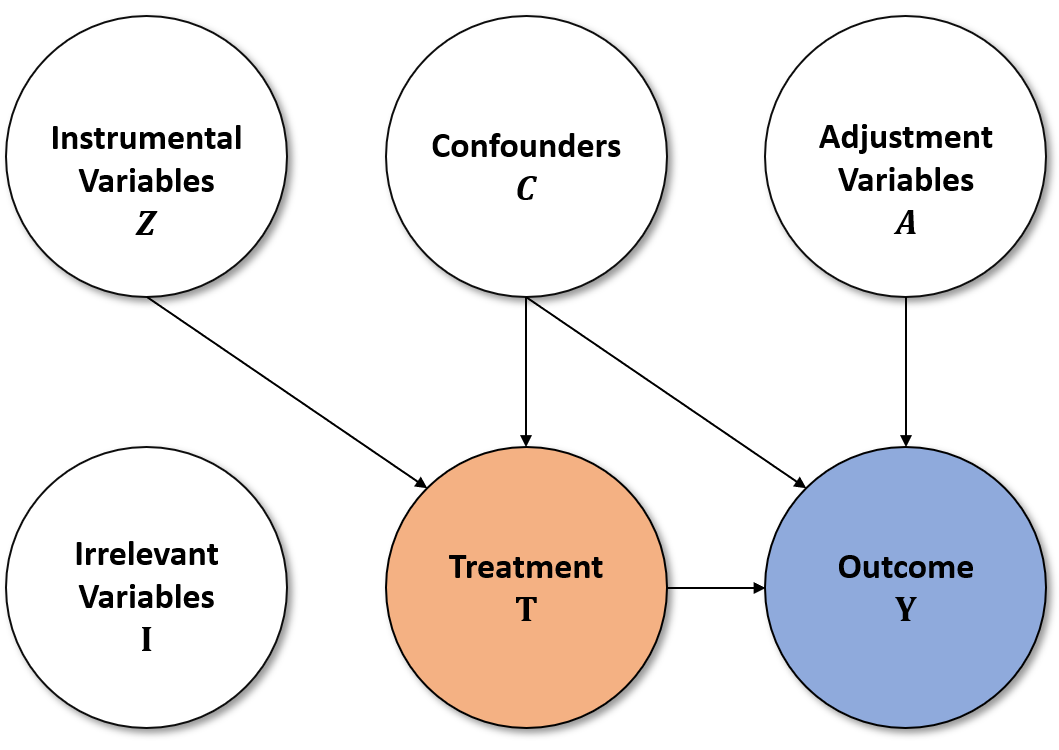}
    \caption{The types of observed variables and the interrelations among these variables.}
    \label{fig: decompose}
\end{figure}

\subsubsection{Synthetic Dataset} To mimic situations where large numbers of variables and information on instrumental, adjustment, confounding, and irrelevant variables are available, we generate a synthetic dataset that reflects the complexity of observational data. Our synthetic data include confounders, instrumental variables, adjustment, and irrelevant variables. The interrelations among these variables, treatments, and outcomes are illustrated in Fig.~\ref{fig: decompose}. The number of observed variables in the vector $X= (C^\intercal,Z^\intercal,I^\intercal,A^\intercal)^\intercal$ is set to 60, including 15 confounders in $C$, 15 adjustment variables in $A$, 10 instrumental variables in $Z$, and 20 irrelevant variables in $I$. The model used to generate the continuous outcome variable $Y$ in this simulation is the partially linear regression model (Eq.~(\ref{Eqn: sim})), extending the ideas described in~\cite{chu2020matching,robinson1988root}: 
\begin{equation}
Y=\tau((C^\intercal, A^\intercal)^\intercal)T + g((C^\intercal, A^\intercal)^\intercal)+ \epsilon, \\
\label{Eqn: sim}
\end{equation}
where $T\overset{ind.}{\thicksim}\text{Bernoulli}(e_0((C^\intercal, Z^\intercal)^\intercal))$. The function $\tau((C^\intercal,A^\intercal)^\intercal)$ describes the true treatment effect as a function of the values of adjustment variables $A$ and confounders $C$. The function $g((C^\intercal, A^\intercal)^\intercal)$ can have an influence on outcome regardless of treatment assignment. 

\vspace{-3mm}
\subsubsection{Results and Analysis} As shown in Table \ref{result3}, our method significantly outperforms competitive baselines, such as TARNET, SITE, CFRNET, and CMGP. These baseline methods all rely on the assumption that all observed variables are pre-treatment variables. However, this assumption is not tenable for observational data in practice, which may include instrumental, adjustment, confounding, and irrelevant variables, as in our simulated dataset. Results in Table \ref{result3} show that our IDRL method can effectively filter out the information about treatment domains and irrelevant information for predicting the potential outcomes, especially without any loss of common predictive information. Besides, we conduct ablation studies and report the performance of our method without the Information Maximization Module or Domain-Independent Module, respectively. From the results, we can see that performance suffers when either is left out, which demonstrates the effectiveness of these two modules in our method.

\begin{table}[t]
  \caption{Performance on simulated dataset. We present the mean value of $\sqrt{\epsilon_\text{PEHE}}$ and $\epsilon_\text{ATE}$ on the test sets.}
  \label{result3}
  \centering
  \scalebox{1}{
  \begin{tabular}{lll}
    \toprule

    Method     & $\sqrt{\epsilon_\text{PEHE}}$   & $\epsilon_\text{ATE}$ \\
    \midrule
    TARNET & 1.36 & 0.31 \\
    SITE & 1.15 & 0.25 \\
    CFRNET & 0.98 & 0.19  \\
    CMGP & 0.83 & 0.17 \\
    \midrule
    IDRL w/o $MI(r_i,s)$ & 0.89 &0.17  \\
    IDRL w/o $MI(r_i,h_i)$ & 1.24 &0.27 \\
    IDRL & 0.62 & 0.13 \\
    \bottomrule
  \end{tabular}
  }
\end{table}

\begin{figure}[t]
    \centering
    \includegraphics[width=0.98\columnwidth]{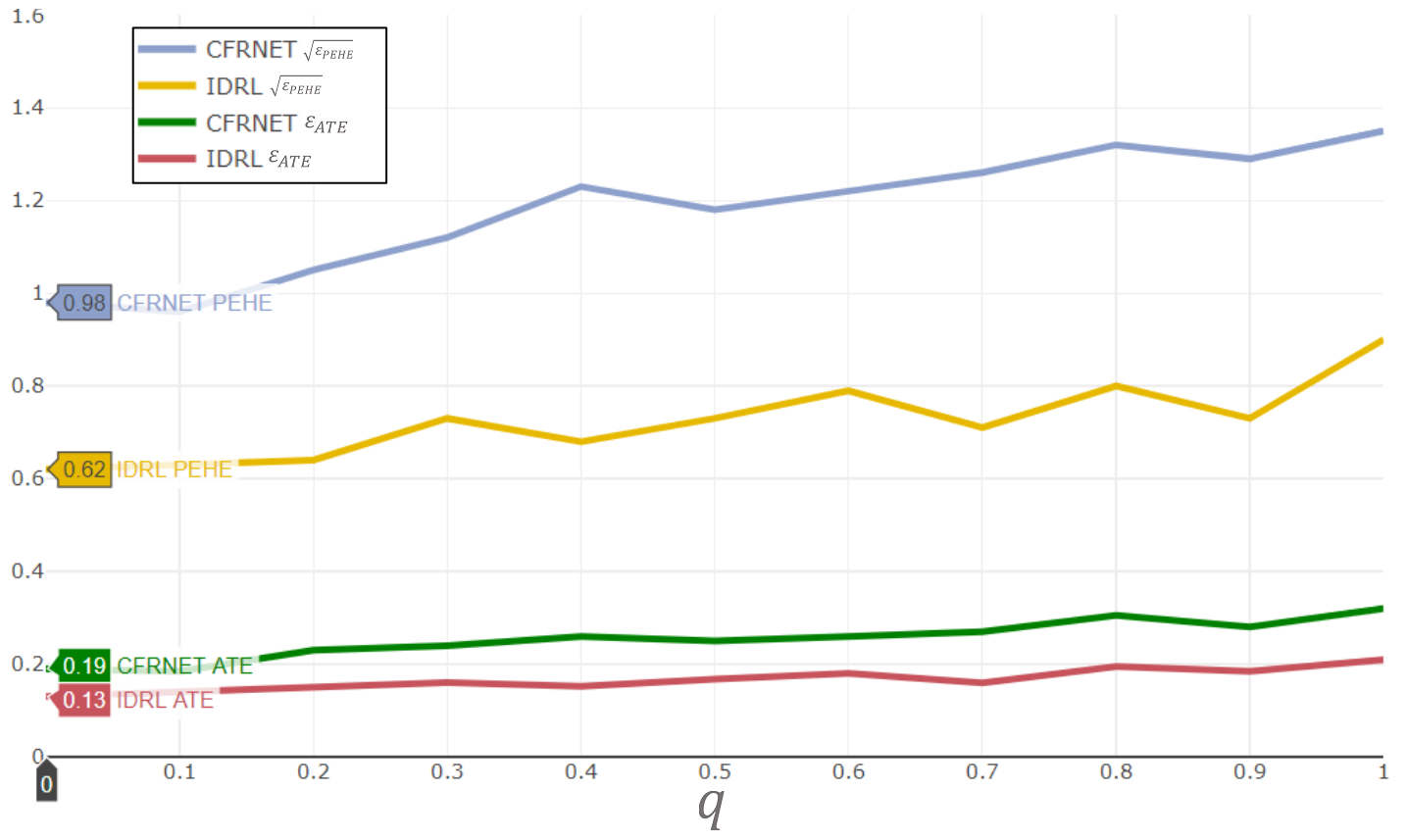}
    \caption{$\sqrt{\epsilon_\text{PEHE}}$ and $\epsilon_\text{ATE} $ performance on simulation dataset with $q$ from 0 to 1.}
    \label{fig: bias}
\end{figure}

In addition, we evaluate the robustness of our proposed method with respect to different levels of treatment selection bias. Although in our simulation procedure, the treatment selection bias has been taken into account based on their own propensity score $e_0((C^\intercal, Z^\intercal)^\intercal)$, we use conditional sampling from treatment and control groups to increase the treatment selection bias. If the propensity score $e_0$ is equal to constant 0.5, it means no matter what the confounders and instrumental variables are, the unit is randomly assigned to either the treatment or the control group with the same probability, so that there is no treatment selection bias. The greater $|e_0((C^\intercal, Z^\intercal)^\intercal)-0.5|$ is, the larger selection bias will end up getting. Following the setting in~\cite{shalit2017estimating}, with probability $1-q$, we randomly draw the treatment and control units; with probability $q$, we draw the treatment and control units that have the greatest $|e_0((C^\intercal, Z^\intercal)^\intercal)-0.5|$. Thus, the higher the $q$ is, the larger the selection bias is. We run CFRNET and our method on the simulation datasets with $q$ from 0 to 1, and show the results in Fig.~\ref{fig: bias}. We can observe that our method consistently outperforms the baseline methods under different levels of divergence and is robust to a high level of treatment bias. 

\begin{figure}[t]
    \centering
    \includegraphics[width=0.95\columnwidth]{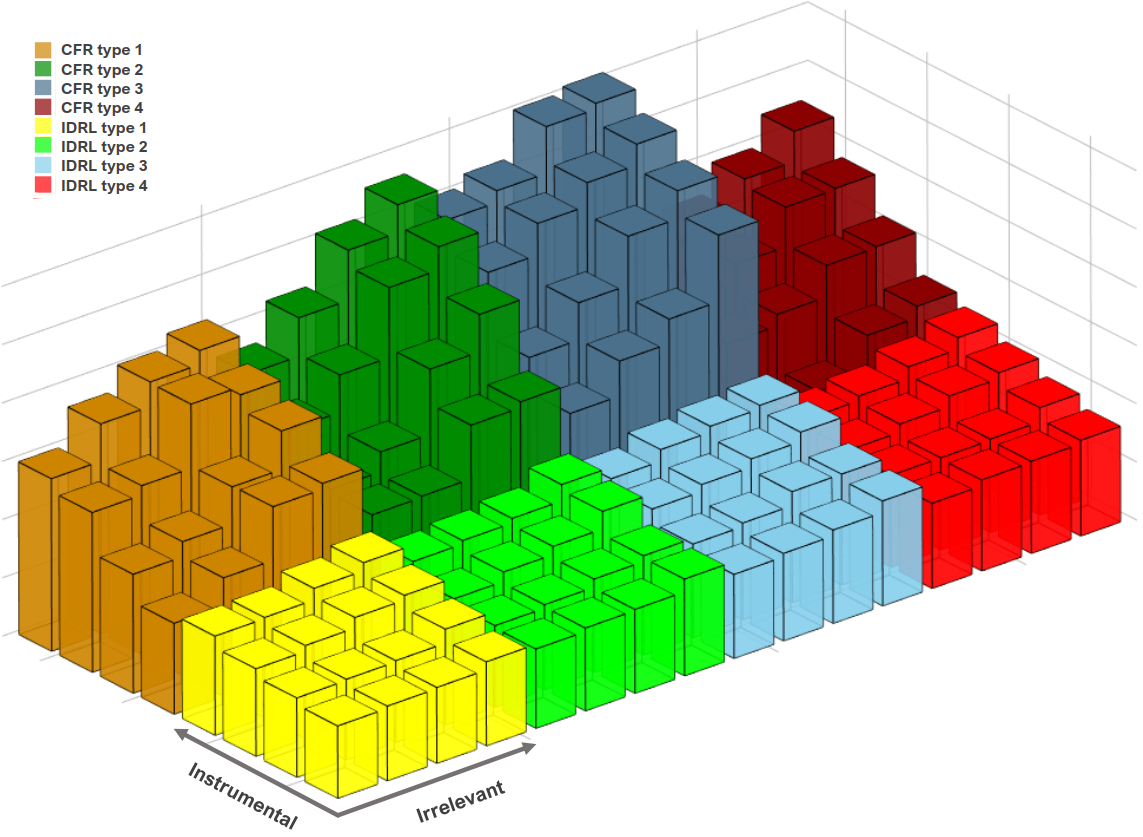}
    \caption{Performance on simulated dataset under different settings. We present the mean value of $\sqrt{\epsilon_\text{PEHE}}$ of CFRNET (Top row) and IDRL (Bottom row). Each color represents one type of domain distribution, such as yellow, green, blue, and red. For each color block, the numbers of the instrumental and irrelevant variables increase, respectively. Our IDRL consistently outperforms the baseline CFRNET under any situation. }
    \label{fig: evaluate}
\end{figure}

To evaluate the sensitivity of our method to different characteristics of the domain distributions and the capability of handling the instrumental and irrelevant variables, we generate four different correlation matrices with different levels of correlation for all observed variables in the simulated dataset. Besides, based on different correlation matrices in the simulation procedure, we gradually increase the numbers of instrumental and irrelevant variables, respectively. As shown in Fig.~\ref{fig: evaluate}, we can observe our IDRL consistently outperforms the baseline CFRNET under any situation. Our method is robust to four types of different domain distributions, and the increase in numbers of instrumental and irrelevant variables, compared to the performance of CFRNET.

\vspace{-3mm}
\section{Related Work}
Learning from observational data requires adjusting for the covariate shift that exists between treatment and control groups~\cite{johansson2016learning,li2017matching,li2016matching_digital_ijcai16,yao2019estimation,cui2020causal}. Balancing neural networks (BNNs)~\cite{johansson2016learning} and counterfactual regression networks (CFRNET)~\cite{shalit2017estimating} are proposed to balance covariate distributions across treatment and control groups by formulating the problem of counterfactual inference as a domain adaptation problem and by enforcing domain invariance with distributional distances such as Wasserstein distance and Maximum Mean Discrepancy. These models may be extended to any number of treatments even with continuous parameters, as described in the perfect match (PM) approach~\cite{schwab2018perfect} and DRNets~\cite{schwab2019learning}. A local similarity preserved individualized treatment effect (SITE) estimation method~\cite{yao2018representation} is proposed to use hard samples to learn representations that preserve local similarity information and balance the data distributions. In~\cite{zhang2020learning}, the authors argue that domain invariance is often too strict a requirement and use counterfactual variance to measure the distributional overlap. In~\cite{li2017matching}, counterfactual prediction is modeled as a classification problem, and matching is conducted based on balanced and nonlinear representations. In~\cite{alaa2017bayesian}, the covariate shift is alleviated via a risk-based empirical Bayes method by minimizing the empirical error in factual outcomes and the uncertainty in counterfactual outcomes.  For our IDRL method, we balance the trade-off between achieving covariate balance and preserving predictive power. In the meanwhile, we leverage mutual information to capture the common predictive information and filter out the information about domains and irrelevant information.

\vspace{-3mm}
\section{Conclusion}
In this paper, we propose the Infomax and Domain-independent Representation Learning (IDRL) method for treatment effect estimation with real-world data. IDRL offers a new thought in probing into the covariate imbalance problem in causal inference.  Circumventing the traditional strategy of enforcing balance between treatment and control groups, IDRL leverages mutual information to address the selection bias and capture the common predictive information. Our method exhibits reliable prediction performances when facing data with different characteristics of data distributions, complicated variable types, and severe covariate imbalance. 

\balance
\vspace{-3mm}
% \bibliographystyle{plain}
% \bibliography{main}

\end{document}